\documentclass[
 amsmath,amssymb,
 aps,
pre,twocolumn]{revtex4-2}
\usepackage{float, xcolor}
\usepackage{wrapfig}
\usepackage{graphicx}
\usepackage{dcolumn}
\usepackage{soul}
\usepackage{geometry}
\usepackage{amssymb}
\usepackage{booktabs}
\usepackage{array} 
\usepackage{cancel}
\usepackage{bm}
\usepackage{subfigure}
\usepackage{ulem} 
\usepackage{threeparttable} 
\usepackage{wrapfig}
\usepackage{xcolor}
\usepackage{multirow}
\usepackage{graphicx}
\usepackage{dcolumn}
\usepackage{ulem}
\usepackage[colorlinks,citecolor=blue,urlcolor=blue,linkcolor=blue,breaklinks=true]{hyperref}
\usepackage[mathlines]{lineno}
\usepackage[marginal]{footmisc}
\newtheorem{property}{Property}
\newtheorem{definition}{Definition}

\begin{document}


\title{Hamiltonian Graph Inference Networks: Joint structure discovery and dynamics prediction for lattice Hamiltonian systems from trajectory data}

\author{Ru Geng \textsuperscript{1} }

\author{Panayotis Kevrekidis \textsuperscript{2}}
 
\author{Yixian Gao\textsuperscript{3}}

\author{Hong-Kun Zhang\textsuperscript{4}}
\email{corresponding author, hongkun@math.umass.edu}

\author{Jian Zu \textsuperscript{3}}
\email{corresponding author, zuj100@nenu.edu.cn}

\address{%
1.School of Mathematics and Statistics, Changchun University, Changchun, 130022, China
}%

\address{%
2.Department of Mathematics and Statistics, University of Massachusetts,   Amherst, 01003, MA, USA,
Department of Physics, University of Massachusetts,   Amherst, 01003, MA, USA,
Department of Mechanical Engineering, Seoul National University, Seoul, Korea,
}%

\address{%
3.Center for Mathematics and Interdisciplinary Sciences,
School of Mathematics and Statistics,  Northeast Normal University,  Changchun, 130024, P.R. China
}%

\address{%
4.School of Sciences, Great Bay University, Dongguan, 523000, PR China
}%

\date{\today}

\begin{abstract}
Lattice Hamiltonian systems underpin models across condensed matter, nonlinear optics, and biophysics, yet learning their dynamics from data is obstructed by two unknowns: the interaction topology and whether node dynamics are homogeneous. Existing graph-based approaches either assume the graph is given or, as in $\alpha$-separable graph Hamiltonian network, infer it only for separable Hamiltonians with homogeneous node dynamics. We introduce the Hamiltonian Graph Inference Network (HGIN), which jointly recovers the interaction graph and predicts long-time trajectories from state data alone, for both separable and non-separable Hamiltonians and under heterogeneous node dynamics. 
HGIN couples a structure-learning module---a learnable weighted adjacency matrix trained under a Hamilton's-equations loss---with a trajectory-prediction module that partitions edges into physically distinct subgraphs via $k$-means clustering, assigning each subgraph its own encoder and thereby breaking the parameter-sharing bottleneck of conventional GNNs. On three benchmarks---a Klein--Gordon lattice with long-range interactions and two discrete nonlinear Schr\"odinger lattices (homogeneous and heterogeneous)---HGIN reduces long-time energy prediction error and trajectory prediction error by six to thirteen orders of magnitude relative to baselines. A symmetry argument on the Hamiltonian loss further shows that the learned weights encode the parity of the underlying pair potential, yielding an interpretable readout of the system's interaction structure.
\end{abstract}

\maketitle


\section{Introduction}\label{sec1}

\begin{figure*}[htp]  %
\centering
\includegraphics[scale=0.4]{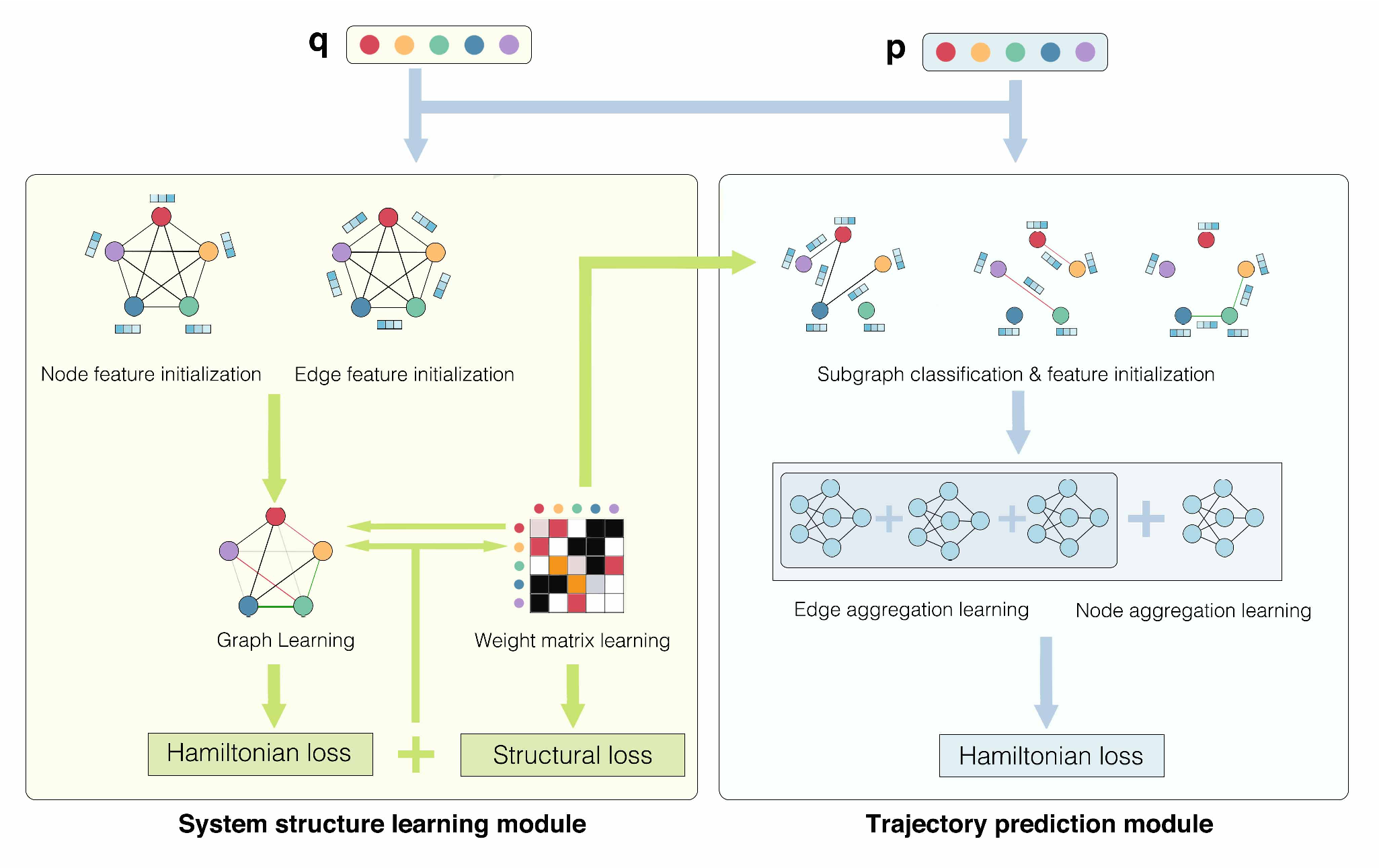}
\caption{Architecture of the Hamiltonian Graph Inference Network (HGIN). The input is the observed particle trajectories $(\mathbf{q},\mathbf{p})$. In the structure-learning module, trajectories are mapped to node features, combined into edge features, and processed by a block whose weighted adjacency $\mathcal{W}_{\theta}$ is randomly initialized and then optimized jointly under a Hamiltonian loss and a Frobenius regularizer. In the trajectory-prediction module, the learned adjacency is partitioned into subgraphs by a $k$-means clustering; see Sec.~\ref{sec:subgraph}. Each subgraph receives its own node and edge encoder, so that the parameter-sharing constraint of standard GNNs is removed. Integrating the resulting Hamiltonian vector field yields the predicted trajectory for any initial condition.}\label{fig_HGIN}
\end{figure*} 

Nonlinear lattice Hamiltonian systems arise across condensed matter physics \cite{lewenstein2007ultracold}, high-energy physics \cite{jiao2015pepx}, cosmology \cite{bruneton2012dynamics}, materials science \cite{zok2019integrating}, chemistry \cite{coe2019lattice}, biology \cite{ramakrishna2023bio}, and medicine \cite{hu2019novel}.
In the era of machine-learning (ML) and associated applications, one can envision novel
opportunities towards learning and identifying Hamiltonian systems (and potential associated
nonlinear as well as wave) phenomena~\cite{PGKreview}. In particular, an intriguing scenario
involves the possibility of possessing either experimental or (large scale) computational
data and seeking ---using ML techniques--- an effective mathematical (Hamiltonian) description thereof~\cite{gao2025alpha,saqlain2022discovering} in analogy to highly popular 
continuum system identification methods such as SINDy~\cite{bpk,rudy2017data} or PINNs~\cite{raissi2019physics,karniadakis2021physics}.
 
Artificial neural networks (ANNs) exhibit unique advantages in multiple aspects: predicting system evolution \cite{david2023symplectic,robinson2022physics,nuske2023finite}, classifying complex patterns \cite{radhakrishnan2023wide,arnold2022replacing,ruiz2023neural}, solving differential equations \cite{cao2024laplace,chen2022deep}, and revealing hidden conserved quantities within systems \cite{lu2023discovering,liu2021machine}. For lattice systems in particular, ANNs have been used for the discovery of governing equations \cite{saqlain2023discovering}, identification of phase transition points \cite{zvyagintseva2022machine}, computation of numerical solutions \cite{zhu2022neural}, identification of conservation laws \cite{chen2025data}, and prediction of motion trajectories \cite{jin2022learning}. Two limitations persist: (i) some methods rely on known equations \cite{zhu2022neural} or require constructing overcomplete operator libraries \cite{saqlain2023discovering}, which restricts their applicability. (ii) Most work focuses on separable Hamiltonian systems (where kinetic and potential energy terms are explicitly decoupled), while non-separable Hamiltonian systems, common in fluid dynamics and quantum mechanics, have received little attention.
 
When the interactions between particles are known, Graph Neural Networks (GNNs) can encode this topology directly: their message-passing mechanism captures the connectivity pattern and represents the resulting interaction structure. GNNs have been applied successfully to classical interacting systems~\cite{atkinson2022improved,bishnoi2023learning,sanchez2019hamiltonian,schuetz2022combinatorial}. Non-separable symplectic neural networks (NSSNN)~\cite{xiong2021nonseparable} extend this idea to non-separable Hamiltonians, but for multi-particle systems they still require the graph to be supplied in advance and assume homogeneous node dynamics---i.e., every node evolves under the same equation---so that parameter sharing across nodes is justified (see Sec.~\ref{sec:basic} for the precise definition).

In many experimental settings, however, the interaction topology is not directly observable, and it is not known a priori whether the node dynamics are homogeneous. This motivates the central question of the present work: \emph{can the interaction graph be inferred solely from trajectory data, without any prior knowledge of the system's topology?}

When the underlying graph is not given, existing GNN approaches can only handle globally coupled, homogeneous systems (where every particle pair interacts under the same rule). Lattices with sparse, non-global coupling require either partial structural knowledge or supervised link labels~\cite{liang2020cryspnet,gu2019link}, which sharply limits the scope of GNNs. Recent work by~\cite{gao2025alpha} proposed $\alpha$-separable graph Hamiltonian network ($\alpha$-SGHN), which autonomously discovers lattice structure from trajectory data, but only under two restrictive assumptions: the Hamiltonian must be separable, and every node must obey the same equation. These assumptions rule out many lattice models of physical interest---notably non-separable nonlinear Schr\"odinger lattices and lattices with  heterogeneous node dynamics---which are the settings we target in this paper.

 \begin{table*}[]
 \centering
\caption{Network performance comparison.  HGIN is the only method requiring no prior interaction topology while covering every scenario and predictive capability listed.}
\label{tab:model_comparison}
\renewcommand{\arraystretch}{1.3} 
\begin{ruledtabular}
\begin{tabular}{lcccccc}
\multicolumn{1}{l}{\multirow{2}{*}{Model}} &Requirement & \multicolumn{2}{c}{Applicable scenarios} & \multicolumn{3}{c}{Predictive ability} \\
\cmidrule(lr){2-2} \cmidrule(lr){3-4} \cmidrule(lr){5-7}
\multicolumn{1}{c}{}   & without IntRel  & NonSepHam                 & HetNodDyn                & IntRel        & IntStr        & EvnSymPotEne        \\
\midrule
MLP                                       & $\checkmark$          &  $\checkmark$                  & $\checkmark$                   & $\times$         & $\times$           &  $\checkmark$                   \\
HNN                                      & $\checkmark$         &  $\checkmark$                  &  $\checkmark$                  & $\times$           & $\times$           &  $\checkmark$               \\
SympNet                                  & $\checkmark$          &  $\checkmark$                  &  $\checkmark$                   & $\times$          & $\times$          &  $\checkmark$                \\
NSSNN                                       & $\times$       &  $\checkmark$                 & $\times$                    & $\times$           & $\times$           &  $\checkmark$              \\
HOGN                                      &  $\times$      & $\checkmark$                 & $\times$                  & $\times$          &$\times$          & $\times$              \\
HGNN                                       &  $\times$      & $\times$                   &$\times$                     &$\times$          & $\times$          & $\times$                 \\
$\alpha$-SGHN                                    &  $\checkmark$          & $\times$                    & $\times$                     &  $\checkmark$        & $\times$          & $\checkmark$                   \\
HGIN (Ours)                                       &  $\checkmark$  &  $\checkmark$                &  $\checkmark$                & $\checkmark$         &  $\checkmark$         &  $\checkmark$   
\\       
\end{tabular}
\end{ruledtabular}
\footnotesize
\text{Note: IntRel: interaction relationship, NonSepHam: non-separable Hamiltonian system, HetNodDyn: Heterogeneous }
 \\
\text{node dynamics,  IntStr: interaction strength, EvnSymPotEne: even-symmetric potential energy}
\end{table*}
 
We propose the Hamiltonian Graph Inference Network (HGIN), which discovers the system structure and predicts trajectories from observational data alone, with no prior topological knowledge. The model comprises two modules (Fig.~\ref{fig_HGIN}):

The structure-learning module (left panel of Fig.~\ref{fig_HGIN}) optimizes a learnable weighted adjacency $\mathcal{W}_\theta$ jointly under a Hamiltonian residual loss and a Frobenius regularizer. At convergence, the entries of $\mathcal{W}_\theta$ encode both the interaction strength between each pair of particles and the parity of the underlying pair potential; see Property~\ref{prop:symmetry}.

The trajectory-prediction module (right panel of Fig.~\ref{fig_HGIN}) partitions the learned $\mathcal{W}_\theta$ into subgraphs by a $k$-means clustering. Each subgraph is assigned its own node and edge encoder, so that physically distinct interactions are represented by distinct networks rather than by a single shared pair $(g_\mathrm{noc}, g_\mathrm{enc})$. Integrating the resulting Hamiltonian vector field produces predicted trajectories.

Table~\ref{tab:model_comparison} compares HGIN with existing methods: multilayer perceptron (MLP) \cite{goodfellow2016deep}, Hamiltonian neural network (HNN) \cite{greydanus2019hamiltonian}, symplectic neural networks (SympNets) \cite{jin2020sympnets}, NSSNN, Hamiltonian ODE graph network (HOGN) \cite{sanchez2019hamiltonian}, Hamiltonian graph neural networks (HGNN) \cite{bishnoi2023learning}, and $\alpha$-SGHN.
HGIN is the only model that neither requires the interaction graph as an input nor imposes separability on the Hamiltonian or homogeneity on the node dynamics.

We summarize our main contributions as follows. 

1) Our model learns the potential interactions among system particles under the assumption that only trajectory data is available. Standard GNN methods require prior knowledge of the graph structure; our approach does not.
﻿
2) Our model applies to both separable and non‑separable Hamiltonian systems.
﻿
3)  Our model can identify whether the system exhibits heterogeneous node dynamics and perform classification accordingly.

The remainder of the paper is organized as follows. Sec.~\ref{sec:basic} introduces the mathematical framework. Sec.~\ref{sec:method} describes HGIN. Sec.~\ref{sec:results} reports numerical results, and Sec.~\ref{sec:conclusions} concludes.

\section{Basic concepts and problem formulation}\label{sec:basic}
\subsection{Basic concepts}

Let $\mathcal{G}=(\mathcal{V},\mathcal{E})$ be a directed graph with $N$ nodes $\mathcal{V}=\{v_1,\ldots,v_N\}$ and directed edges $\mathcal{E}=\{e_{i,j}\mid j\in\mathcal{N}_i\}$, where $\mathcal{N}_i$ is the out-neighbor set of $v_i$. We represent $\mathcal{G}$ by its $N\times N$ weighted adjacency matrix $\mathcal{W}=(w_{ij})$; the entry $w_{ij}$ encodes the influence from $v_j$ to $v_i$. If $w_{ij}=w_{ji}$ for all $i,j$, the graph is undirected.

A lattice system can naturally be viewed as such a graph. Let $x_i\in\mathbb{R}^{n_i}$ denote the state of node $v_i$. We write the dynamics in the general form
\[
\dot{x}_i = f_i(x_i) + \sum_{j\in\mathcal{N}_i} g_{ij}(x_i, x_j), \quad i = 1,\dots,N,
\]
where $f_i\colon\mathbb{R}^{n_i}\to\mathbb{R}^{n_i}$ is the intrinsic dynamics of node $i$ and $g_{ij}\colon\mathbb{R}^{n_i}\times\mathbb{R}^{n_j}\to\mathbb{R}^{n_i}$ is the coupling from $v_j$ to $v_i$.

\begin{definition}\label{def:hom}
We say the system has \emph{homogeneous node dynamics (HomNodDyn)} if all nodes share the same intrinsic and coupling functions:
\[
\dot{x}_i = f(x_i) + \sum_{j\in \mathcal{N}_i} g(x_i, x_j), \quad i = 1,\dots,N.
\]
\end{definition}

\begin{definition}\label{def:het}
We say the system has \emph{heterogeneous node dynamics (HetNodDyn)} if at least one of the following holds.

Case~1: all nodes follow the same functional forms $f$ and $g$, but with node-dependent parameters $\theta_i$, $\psi_i$:
\[
\dot{x}_i = f(x_i;\theta_i) + \sum_{j\in \mathcal{N}_i} g(x_i, x_j;\psi_i), \quad i = 1,\dots,N.
\]

Case~2: the functional forms themselves differ across nodes. More precisely, there exist at least two nodes with distinct intrinsic dynamics $f_i\neq f_j$, or at least two coupling functions $g_{ij}\neq g_{ik}$:
\[
\dot{x}_i = f_i(x_i) + \sum_{j\in \mathcal{N}_i} g_{ij}(x_i, x_j), \quad i = 1,\dots,N.
\]
\end{definition}

\begin{definition}\label{def:even}
We say a coupling function $g$ has \emph{even symmetry} if $g(x_i,x_j)=g(x_j,x_i)$ for all admissible $(x_i,x_j)$.
\end{definition}

\subsection{Problem formulation}

Given observed trajectory samples $\{(\mathbf{q}^t,\mathbf{p}^t)\}_{t=t_1}^{t_T}$ from a lattice Hamiltonian system, we aim to: (i) recover the underlying interaction graph; (ii) decide whether the interactions are homogeneous or heterogeneous; and (iii) learn a Hamiltonian that reproduces the observed dynamics and predicts trajectories from unseen initial conditions.
 
\section{Methodology}\label{sec:method}
\subsection{Structure-learning module}

We represent the lattice as a directed graph $\mathcal{G}=(\mathcal{V},\mathcal{E})$ in which node $v_i$ encodes particle $\alpha_i$ and every ordered pair of distinct nodes carries an edge whose weight $w_{ij}$ measures the interaction strength from $\alpha_j$ to $\alpha_i$. Because the edge set is unknown a priori, we initialize $\mathcal{W}_{\theta}=(w_{ij})$ at random and let the training procedure decide which weights should vanish.

The node feature of particle $\alpha_i$ at time $t$ is
\begin{eqnarray}\label{n_t}
	\mathbf{n}_{i}^{(t)} = \left[q^t_i-p^t_i, q^t_i  \odot p^t_i \right],
\end{eqnarray}
where $\odot$ is the elementwise (Hadamard) product. The first component records the phase-space imbalance between position and momentum at the node, while the second captures their nonlinear coupling; together they give the network layers downstream enough freedom to approximate non-separable Hamiltonians.

The edge feature encodes the relative phase-space state of the two incident particles,
\begin{eqnarray}\label{eij_t}
	\mathbf{e}_{ij}^{(t)} = \left[\mathbf{h}_j^{(t)} - \mathbf{h}_i^{(t)}, \mathbf{h}_j^{(t)} \odot \mathbf{h}_i^{(t)}\right],
\end{eqnarray}
where $\mathbf{h}^t_i=(q^t_i,p^t_i)$. The displacement term $\mathbf{h}_j^{(t)}-\mathbf{h}_i^{(t)}$ captures phase-space differences, and the product term $\mathbf{h}_j^{(t)}\odot\mathbf{h}_i^{(t)}$ captures elementwise correlations; the combination lets a downstream network represent non-separable pair interactions.

We then parameterize the system Hamiltonian as
\begin{eqnarray}
\mathcal{H}^{(t)} = \sum_{i\in\mathcal{V}} g_{\mathrm{noc}}(\mathbf{n}_{i}^{(t)}) + \sum_{i,j\in\mathcal{V}} w_{ij}\, g_{\mathrm{enc}}(\mathbf{e}_{ij}^{(t)}),
\end{eqnarray}
where $g_{\mathrm{noc}}$ and $g_{\mathrm{enc}}$ are MLP encoders for node and edge features, and $w_{ij}$ are the learnable graph weights.

Structure learning and Hamiltonian fitting are coupled through a single multi-task loss
\begin{eqnarray}
\mathcal{L} = \frac{1}{T} \sum_{t=t_1}^{t_T} \mathcal{L}_{\text{phy}}^{(t)} + \gamma \mathcal{L}_{\text{GL}},
\end{eqnarray}
with $\mathcal{L}_{\mathrm{GL}} = \|\mathcal{W}_{\theta}\|_F^2$ a Frobenius regularizer on the graph weights, $\mathcal{L}_{\mathrm{phy}}^{(t)} = \delta(\mathcal{H}^{(t)})$ the residual of Hamilton's equations (defined below), and $\gamma\in[0,1]$ a scalar trade-off. The Hamiltonian residual is

\begin{eqnarray}\label{lossphy}
\delta (\mathcal{H}^{(t)}) = \left\| \frac{\partial \mathcal{H}^{(t)}}{\partial \mathbf{p}^t} - \mathbf{\dot{q}}^t \right\|_2^2 + \left\| -\frac{\partial \mathcal{H}^{(t)}}{\partial \mathbf{q}^t} - \mathbf{\dot{p}}^t \right\|_2^2.
\end{eqnarray}

\subsection{Trajectory prediction module}

\subsubsection{Subgraph classification}\label{sec:subgraph}
The structural learning module outputs a weighted adjacency matrix \( \mathcal{W}_{\theta} = (w_{ij}) \), where each entry \( w_{ij} \in \mathbb{R} \) quantifies the interaction strength between particles \( \alpha_i \) and \( \alpha_j \). 
The weighted adjacency matrix  has the following empirically observed property:
\begin{property}\label{prop:symmetry}
If the interaction potential between particles $\alpha_i$ and $\alpha_j$ satisfies $V_{ij}(q_i,q_j) = V_{ij}(q_j,q_i)$ (i.e., is symmetric under particle exchange), then by the symmetry of the Hamiltonian loss, the learned weights satisfy $w_{ij}\approx w_{ji}$ at convergence. Conversely, if the learned weights are asymmetric ($w_{ij}\neq w_{ji}$), this signals a non-even interaction potential.
\end{property}
 
The optimal number of clusters is determined using the silhouette coefficient method, and the k-means algorithm is used to group similar matrix values together.

We apply the $k$-means algorithm  to cluster the learned weighted adjacency matrix and use the silhouette coefficient method \cite{sai2017optimal} to determine the optimal number of clusters, denoted as \(K+1\). Accordingly, the matrix elements can be partitioned into \(K\) effective categories and one noise class; see also the caption of Fig.~\ref{fig_kglri_edges_class}.
Specifically, the clusters are sorted in ascending order of their mean values and denoted as \(\mathcal{C}_0, \mathcal{C}_1, \dots, \mathcal{C}_K\), where \(\mathcal{C}_0\) corresponds to weights close to zero resulting from machine error, indicating the absence of an edge at the corresponding position. The remaining clusters \(\mathcal{C}_1, \dots, \mathcal{C}_K\) form \(K\) subgraphs. Elements within each cluster share identical or similar weight values, suggesting that the interaction strengths between the corresponding particles are consistent and can therefore be described by the same function.

To distinguish between self-interactions and interactions between distinct particles, we further divide \(\mathcal{C}_1, \dots, \mathcal{C}_K\) into two groups: one group consists of elements located on the diagonal positions of the matrix, and the number of such clusters is denoted as \(u\); the other group consists of elements located on the off-diagonal positions, with the number of clusters denoted as \(v\), satisfying \(u+v=K\). In particular, if two weights located at symmetric positions with respect to the diagonal (i.e., $w_{ij}$ and $w_{ji}$) fall into the same off-diagonal cluster, this indicates that the corresponding interaction function is even. Conversely, if they fall into different clusters, the interaction is non-even. For ease of subsequent description, the off-diagonal clusters are relabeled as \(\mathcal{C}^o_1, \dots, \mathcal{C}^o_v\), and the diagonal clusters as \(\mathcal{C}^d_1, \dots, \mathcal{C}^d_u\).

\subsubsection{Per-subgraph Hamiltonian}

Reusing the node and edge features~\eqref{n_t} and~\eqref{eij_t}, we assign each off-diagonal cluster $\mathcal{C}^o_k$ its own edge encoder $g^k_{\mathrm{enc}}$ and each diagonal cluster $\mathcal{C}^d_k$ its own node encoder $g^k_{\mathrm{noc}}$, writing
\begin{eqnarray}
\mathcal{H}^{(t)}_1 &=& \sum_{k=1}^{v}\sum_{(i,j) \in \mathcal{C}^o_k} w_{ij}\, g^{k}_{\mathrm{enc}}(\mathbf{e}_{ij}^{(t)}), \\
\mathcal{H}^{(t)}_2 &=& \sum_{k=1}^{u}\sum_{i \in \mathcal{C}^d_k}  g^k_{\mathrm{noc}}(\mathbf{n}_{i}^{(t)}), \label{h2t}\\
\mathcal{H}^{(t)}  &=& \mathcal{H}^{(t)}_1 + \mathcal{H}^{(t)}_2.
\end{eqnarray}
Distinct encoders for distinct clusters circumvent the shared-parameter constraint of standard GNNs, which cannot represent heterogeneous node dynamics with a single pair $(g_{\mathrm{noc}},g_{\mathrm{enc}})$.

The trajectory-prediction module is trained by minimizing the Hamiltonian residual
\begin{eqnarray}\label{phy_loss}
\mathcal{L} = \frac{1}{T}\sum_{t=t_1}^{t_T} \mathcal{L}_{\mathrm{phy}}^{(t)},
\end{eqnarray}
where $\mathcal{L}_{\mathrm{phy}}^{(t)}=\delta(\mathcal{H}^{(t)})$ is the same Hamilton's-equations residual of Eq.~\eqref{lossphy}. We train the two modules sequentially: the structure-learning module first runs to convergence, its output $\mathcal{W}_{\theta}$ is frozen, the $k$-means clustering is applied, and the trajectory-prediction module is then trained on the resulting subgraphs with the graph fixed.

\subsubsection{Trajectory prediction}
Once trained, the learned Hamiltonian defines a vector field through Hamilton's equations; integrating this field from any initial condition $(\mathbf{q}_0,\mathbf{p}_0)$ yields the predicted trajectory
\begin{equation}
(\hat{\mathbf{q}}_t,\hat{\mathbf{p}}_t) = (\mathbf{q}_0,\mathbf{p}_0) + \int_{t_0}^{t}\left(\frac{\partial\mathcal{H}}{\partial\mathbf{p}},\,-\frac{\partial\mathcal{H}}{\partial\mathbf{q}}\right)\mathrm{d}s.
\end{equation}

\section{Computational Results}\label{sec:results}

We evaluate HGIN on three lattice Hamiltonian benchmarks spanning the separable and non-separable regimes, and the homogeneous and heterogeneous node-dynamics regimes.

\textbf{Baselines.} Because only trajectories are observed and no graph is supplied, we restrict the baselines to models that do not require a pre-specified interaction topology: MLP, HNN, SympNet (G-type and LA-type), NSSNN, and $\alpha$-SGHN. Graph-based methods such as HOGN and HGNN are excluded from the comparison because they require the graph as an input.
  Across all test benchmarks HGIN uses a single fixed network parameters, whereas the baselines use per-benchmark optimal hyperparameters determined by grid search ({detals see Appendix~B).

We generate 50 trajectories as training data for each test system. Each trajectory spans the interval $[0,10]$\,s and is sampled at $\Delta t=0.05$\,s, giving $10{,}000$ training points in total. Apart from this, we set 30 trajectories as a validation set for grid-search tuning of the baseline hyperparameters, and 30 additional trajectories as the test set. To evaluate long-term prediction we extend the test horizon to $[0,20]$\,s, twice the training duration. Details of the trajectory generation and network configuration are given in the appendix.

We measure performance by the mean squared error
\begin{eqnarray*}
\mathrm{MSE}=\frac{1}{M}\sum_{m=1}^{M}\frac{1}{T}\sum_{t=1}^{T}\frac{1}{N}\sum_{i=1}^{N}\bigl(X_{i,t}^{(m)}-\hat{X}_{i,t}^{(m)}\bigr)^2,
\end{eqnarray*}
where $M$ is the number of test trajectories, $T$ the number of time steps, $N$ the number of particles, $X$ denotes the ground-truth state, and $\hat{X}$ the predicted state.

\subsection{Separable Hamiltonian lattice with homogeneous node dynamics} \label{sec:ex1}

\begin{figure*}[htp]  %
\centering
\includegraphics[scale=0.9]{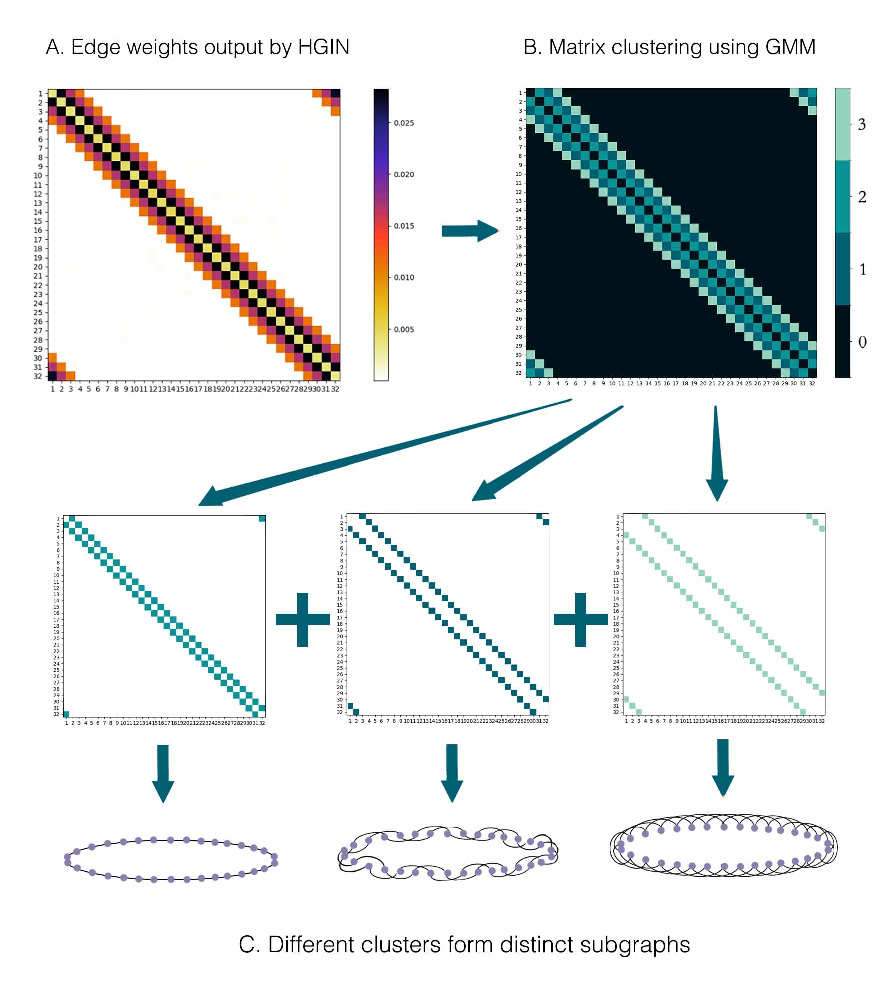}
\caption{KG-LRI structure recovery. (A) Weighted adjacency matrix $\mathcal{W}_{\theta}$ output by the HGIN structure-learning module. (B) $k$-means partition of $\mathcal{W}_{\theta}$ into four clusters $\mathcal{C}_0,\dots,\mathcal{C}_3$, ordered by mean weight. $\mathcal{C}_0$ (value $0$ on the colorbar) collects near-zero entries and is pruned as noise; $\mathcal{C}_1,\mathcal{C}_2,\mathcal{C}_3$ are the three off-diagonal clusters corresponding to nearest-, next-nearest-, and third-nearest-neighbor couplings; 
diagonal entries are classified into one class.
(C) Each off-diagonal cluster is displayed as its own subgraph.}\label{fig_kglri_edges_class}
\end{figure*} 

Our first test example consists of 
 the Klein-Gordon lattice system with long-range interactions (KG-LRI), a standard model for nonlinear dynamics including localized excitations in ionic crystals and related settings~\cite{Flach2008}. The KG lattice system, inclusive of second and third-order long-range interactions, has the following Hamiltonian \cite{pgkreview2}:

\begin{eqnarray}\label{kglri_h}
 H&=&\sum_{i=1}^{32}\big(\frac{1}{2}p_i^2+\frac{1}{2}q_i^2+\frac{1}{4}q_i^4+\frac{1}{4}(q_i-q_{i+1})^2\nonumber \\ 
 &+&\frac{3}{20}(q_i-q_{i+2})^2+\frac{1}{10}(q_i-q_{i+3})^2\big).
 \end{eqnarray}

Note that KG-LRI is a separable Hamiltonian lattice with homogeneous node dynamics: every site obeys the same equation of motion, but the coupling strengths to the nearest-, next-nearest-, and third-nearest-neighbors differ. 

Applied to $\mathcal{W}_{\theta}$ (Fig.~\ref{fig_kglri_edges_class}\,A), the $k$-means method selects four clusters via the silhouette coefficient, shown in Fig.~\ref{fig_kglri_edges_class}\,B. Each cluster corresponds to a distinct term of the Hamiltonian~\eqref{kglri_h}. The near-zero cluster $\mathcal{C}_0$ collects weights that are indistinguishable from machine precision and is pruned as absent edges. Three off-diagonal clusters $\mathcal{C}_1,\mathcal{C}_2,\mathcal{C}_3$ collect the nearest-, next-nearest-, and third-nearest-neighbor couplings respectively; one can check that their mean weights scale in the ratio $0.25:0.15:0.10$, matching the coefficients of the three coupling terms in Eq.~\eqref{kglri_h}. The diagonal entries is classified into one class and absorb the on-site kinetic and potential terms $\tfrac12 p^2+\tfrac12 q^2+\tfrac14 q^4$ into a single node-encoder cluster. Thus the four non-noise clusters reproduce the four distinct functional forms of the true Hamiltonian without any prior knowledge of the couplings.
Note that the diagonal entries are classified into the class close to zero, but they are not actually zero, because part of their coefficients is included in \eqref{h2t}.

Table~\ref{tab:mse_kglri} reports training, test, and long-time energy errors for HGIN and the five baselines.
All models reach low training loss (column~1), with HGIN the lowest. Among the baselines, only HNN generalizes well: its test loss (column~2) stays within the same order as its training loss.  However, its performance is inferior to that of HGIN. For long-term prediction (column~3, energy MSE in 20s), HGIN's error remains orders of magnitude below all baselines and does not grow appreciably with time.

 \begin{table*}[]
 \centering
\caption{KG-LRI benchmark: train loss, test loss, and energy MSE in 20s (twice the training horizon), averaged over 30 trajectories. HGIN outperforms every baseline by at least three orders of magnitude in all three metrics.}
\label{tab:mse_kglri}
\renewcommand{\arraystretch}{1.3} %
\begin{ruledtabular}
\begin{tabular}{lccc }
\multicolumn{1}{c}{}   & Train loss  & Test loss & Energy MSE                         \\
\midrule
MLP  & 2.50$\times10^{-5} $ &  1.51$\times10^{-3}$  & 3.18$\times10^{-1}$  \\
HNN & 2.81$\times10^{-5}$&  9.34$\times10^{-5}$ & 1.30$\times10^{-3}$  \\  
NSSNN & 1.01$\times10^{-4}$  &  2.10$\times10^{+1}$ &  8.15$\times10^{+8}$ \\        
G-SympNet & 6.67$\times10^{-4}$  &  2.59$\times10^{-1}$  &  1.80$\times10^{+0}$  \\ 
LA-SympNet & 6.94$\times10^{-4}$  &  1.48$\times10^{-1}$ &  1.96$\times10^{-1}$ \\ 
$\alpha$-SGHN & 2.42$\times10^{-4}$   &  5.72$\times10^{-4}$   &   1.04$\times10^{-1}$  \\ 
  HGIN (Ours) & 
  $\mathbf{ 2.03 \times 10^{-8}}$  & 
  $\mathbf{ 2.39 \times 10^{-8}}$  &  
  $\mathbf{ 1.53 \times 10^{-7}}$ \\     
\end{tabular}
\end{ruledtabular}
\end{table*}

Fig.~\ref{fig_kglri_state}\,A tracks the time evolution of the  MSE of predicted trajectories averaged over the 30 test trajectories. The baseline errors grow monotonically over the $[0,20]$\,s window, whereas the HGIN curve remains flat to within the log-scale plotting precision.

\begin{figure*}[htp]  %
\centering
\includegraphics[scale=0.9]{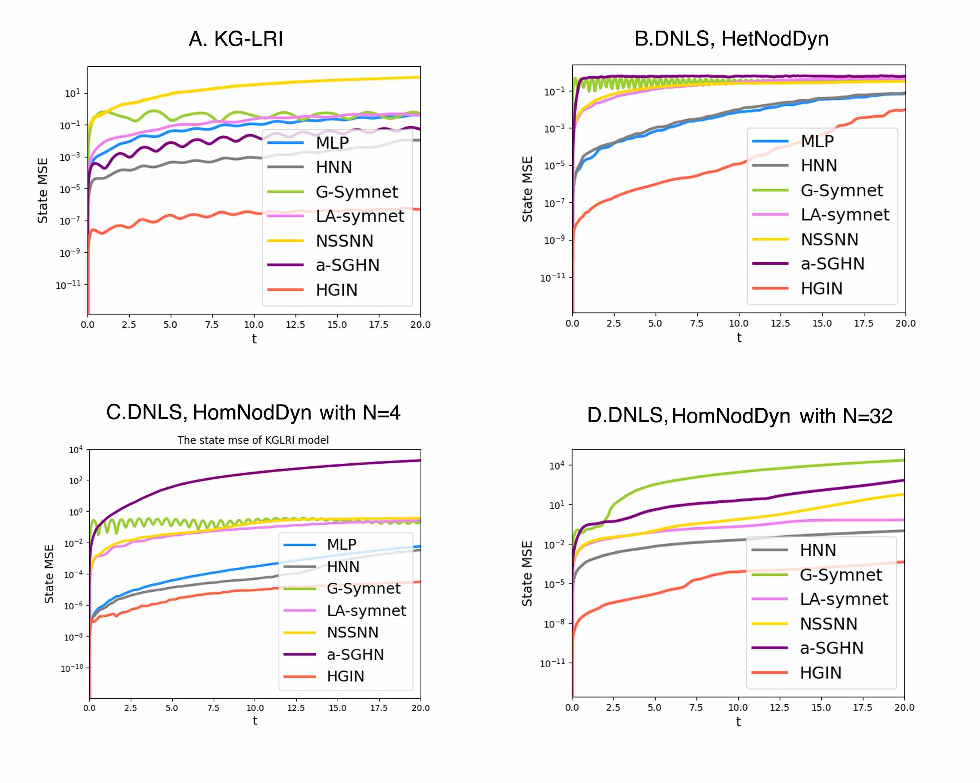}
\caption{Average MSE of predicted trajectories over time on 30 test samples (y-axis on a logarithmic scale). 
Panel A: KG-LRI (Sec.~\ref{sec:ex1}). 
Panel B: Heterogenous DNLS (Sec.~\ref{sec:ex3}). 
Panels C and D: homogeneous DNLS with N=4 and N=32 (Sec.~\ref{sec:ex2}). 
In all cases, HGIN (red) remains essentially flat over the 20-second time window, while each baseline drifts.
}\label{fig_kglri_state}
\end{figure*} 


\subsection{Non-separable Hamiltonian lattice with homogeneous node dynamics} \label{sec:ex2}

\begin{figure*}[htp]  %
\centering
\includegraphics[scale=0.6]{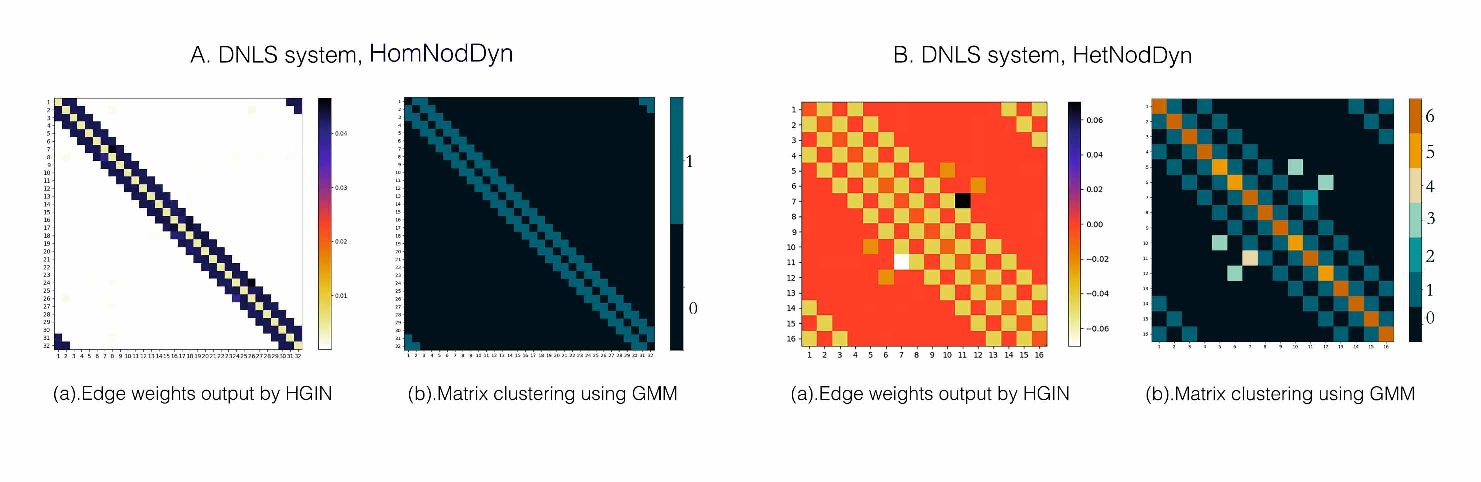}
\caption{Learned weighted adjacency matrices and their $k$-means partitions for the DNLS benchmarks. Panel~A: homogeneous DNLS with nearest- and next-nearest-neighbor couplings [Eq.~\eqref{dnls_h}]---(a) HGIN output, (b) three-cluster $k$-means partition. Panel~B: heterogeneous DNLS with added long-range and cubic terms [Eq.~\eqref{dnls_h2}]---(a) HGIN output, (b) five-cluster $k$-means partition. Colorbar value $0$ denotes edges pruned as machine-precision noise. }\label{fig_dense_class}
\end{figure*} 

The discrete nonlinear Schr\"odinger (DNLS) equation is a prototypical model for nonlinear excitations in lattice systems~\cite{chriseil,kevrekidis2009discrete}. It was originally proposed for electronic transport in biomolecules and has since been applied to nonlinear wave propagation in waveguide arrays and to ultra-cold atom gases in optical lattices~\cite{SzameitRechtsman2024DiscreteNonlinearTopologicalPhotonic,elmar}. We consider a DNLS system with nearest- and next-nearest-neighbor couplings, inspired by~\cite{efremidis2002discrete}, whose Hamiltonian reads
 
  \begin{eqnarray}\label{H1}
 H=\sum_{n=1}^{32} \bigg[ \vert u_n-u_{n-1}\vert^2 + \vert u_n-u_{n-2}\vert^2 -\frac{1}{2}\vert u_n\vert^4  \bigg].\nonumber\\
 \end{eqnarray}
 
Expressing $u_n=p_n+iq_n$ in terms of real and imaginary parts, the Hamiltonian becomes:
 
  \begin{eqnarray}\label{dnls_h}
 H&=&\sum_{n=1}^{32} \bigg[ (p_n-p_{n-1})^2+(q_n-q_{n-1})^2 \nonumber \\
 &+&  \bigg((p_n-p_{n-2})^2+(q_n-q_{n-2})^2 \bigg)\nonumber \\
 &-&\frac{1}{2}(p_n^2+q_n^2)^2\bigg]
 \end{eqnarray}
 
 This example tests a non-separable Hamiltonian lattice system with homogeneous node dynamics. 
Fig. \ref{fig_dense_class} A (a) displays the weighted adjacency matrix learned by HGIN. 
The matrix is partitioned into two clusters, as shown in Fig. \ref{fig_dense_class} A (b), where different colors on the colorbar denote distinct clusters, with 0 indicating the absence of an edge. 
The classification results indicate that the potential and kinetic energy terms of the system, each corresponds to a single function type, which is consistent with the assumptions of the system equations (see Eq.~\eqref{dnls_h}) used to generate the trajectories.

A comparison of the prediction performance between HGIN and the baseline models is presented in Table \ref{tab:DNLS_comparison1}.  
At $N=4$, both HNN and HGIN generalize well (training and test loss are of comparable magnitude), with HGIN achieving the lowest error. At $N=32$, HGIN retains this advantage while the baselines degrade.  

For long-term prediction (energy MSE), HGIN outperforms all baselines by several orders of magnitude. When $N$ increases from 4 to 32 with network parameters held fixed, HGIN's performance remains stable, while the baselines deteriorate.

 \begin{table*}[]
 \centering
\caption{DNLS benchmark with homogeneous node dynamics: train loss, test loss, and energy MSE in 20\,s, averaged over 30 trajectories, for $N=4$ and $N=32$.}
\label{tab:DNLS_comparison1}
\renewcommand{\arraystretch}{1.3} 
\begin{ruledtabular}
\begin{tabular}{lccc|ccc}
\multicolumn{1}{l}{\multirow{2}{*}{Model}} & \multicolumn{3}{c}{N=4} & \multicolumn{3}{c}{N=32} \\
\cmidrule(lr){2-4} \cmidrule(lr){5-7}
\multicolumn{1}{c}{}   & Train loss  & Test loss & Energy MSE  & Train loss  & Test loss & Energy MSE                        \\
\midrule
MLP  &  2.41$\times10^{-6} $ &   1.95$\times10^{-6}$  &  8.29$\times10^{-3}$ &    1.04$\times10^{-4}$  &  6.18$\times10^{-4}$  &    7.30$\times10^{+7}$  \\
HNN &  1.21$\times10^{-6}$&   1.34$\times10^{-6}$ &  6.23$\times10^{-6}$ &   2.65$\times10^{-4}$  &    5.64$\times10^{-4}$  &     6.74$\times10^{-3}$\\  
NSSNN &  1.34$\times10^{-3}$  &  9.80$\times10^{-2}$ &  1.10$\times10^{-2}$ &   2.06$\times10^{-3}$  &   3.28$\times10^{+0}$  &     5.37$\times10^{+8}$\\        
G-SympNet &  6.11$\times10^{-4}$  &  1.17$\times10^{-1}$  &  2.91$\times10^{-3}$ &  1.39$\times10^{-3}$  &  2.66$\times10^{-1}$  &     7.09$\times10^{+20}$\\ 
LA-SympNet &  7.96$\times10^{-4}$  &   1.19$\times10^{-1}$ &   7.55 $\times10^{-3}$ &   9.72$\times10^{-4}$  &    1.51$\times10^{-1}$  &    1.65$\times10^{+00}$\\ 
$\alpha$-SGHN & 1.77$\times10^{-1}$  &  1.51$\times10^{-1}$   &  3.11$\times10^{+13}$ &1.30$\times10^{-1}$ & 1.25$\times10^{-1}$  &2.78$\times10^{+18}$ \\ 
  HGIN (Ours) &  $\mathbf{8.44\times10^{-7}}$  &  $\mathbf{ 7.52\times10^{-7}}$ &  $\mathbf{ 7.23\times10^{-7}}$ &  $\mathbf{ 1.02 \times10^{-7}}$  &   $ \mathbf{2.93\times10^{-7}}
  $  &  $ \mathbf{4.41 \times10^{-6}}$ \\             
\end{tabular}
\end{ruledtabular}
\end{table*}

Fig. \ref{fig_kglri_state} C (N=4) and D (N=32) examine the variation of trajectory prediction MSE over time for the baseline model and HGIN, respectively. HGIN outperforms all baselines.


\subsection{Non-separable Hamiltonian lattice with heterogeneous node dynamics}\label{sec:ex3}

We now turn to the most challenging setting: a non-separable DNLS system with heterogeneous node dynamics. We start from the standard DNLS Hamiltonian~\cite{kevrekidis2009discrete}:
  \begin{eqnarray}\label{H3}
 H&=&\sum_{n=1}^{16}\bigg[  u_nu^{*}_{n+1} +u_n^{*}u_{n+1} +u_nu^{*}_{n+3} \nonumber \\
 &+&u_n^{*}u_{n+3} +\vert u_n\vert^4  \bigg],
 \end{eqnarray}
 where $u_n=p_n+i q_n$, and $u^{*}$ represents the complex conjugate of $u$.
 To test the models on heterogeneous dynamics, we add extra interaction terms. The modified Hamiltonian reads:
 
  \begin{eqnarray}\label{dnls_h2}
 H&=&\sum_{n=1}^{16} \bigg[ 2(p_np_{n+1}+q_nq_{n+1}+p_np_{n+3}+q_nq_{n+3})\nonumber \\
 &+&(p_n^2+q_n^2)^2\bigg]+p_5p_{10}+q_5q_{10}+p_6p_{12}+q_6q_{12}\nonumber  \\
 &+&\frac{1}{3}(q_{11}-q_{7})^3
 \end{eqnarray}

Fig.~\ref{fig_dense_class} B (a) presents the system’s weighted adjacency matrix output by HGIN, along with the weighted adjacency matrix  clustering by $k$-means (Fig.~\ref{fig_dense_class} B (b)). 
According to subfigure B (b), 
each particle in the system has both first-order and third-order neighbors, which belong to the same class, indicating that the interaction forms within this class are identical (consistent with the equation \eqref{dnls_h2} where \(  2(p_np_{n+1}+q_nq_{n+1}+p_np_{n+3}+q_nq_{n+3}) \)).
 The weights between particle pairs ($\alpha_5$, $\alpha_{10}$) and  ($\alpha_6$, $\alpha_{12}$) are grouped into a single class (corresponding to the case where \(  p_5p_{10}+q_5q_{10}+p_6p_{12}+q_6q_{12} \) in the equation \eqref{dnls_h2}).
 The weight between the particle pair ($\alpha_7$, $\alpha_{11}$) is classified separately and is not symmetric, indicating that the potential energy there is non-even symmetric (corresponding to the part of the equation where \( \frac{1}{3}(q_{11}-q_{7})^3 \)).
The diagonal elements are divided into two classes, which corresponds to a violation of the condition that the system should belong to a single class. However, this does not affect subsequent learning, because using two networks to learn the same function has the same effect as using a single network to learn it — a point that will be verified in the subsequent prediction part. This demonstrates a certain robustness of the method.

 \begin{table*}[]
 \centering
\caption{DNLS benchmark with heterogeneous node dynamics: train loss, test loss, and energy MSE over 20\,s, averaged over 30 trajectories.}
\label{tab:model_DNLS2}
\renewcommand{\arraystretch}{1.3} 
\begin{ruledtabular}
\begin{tabular}{lccc }
\multicolumn{1}{c}{}   & Train loss  & Test loss & Energy MSE                         \\
\midrule
MLP  & 3.26$\times10^{-6} $ &  6.61$\times10^{-5}$  & 1.68$\times10^{-2}$  \\
HNN & 1.93$\times10^{-5}$&  1.13$\times10^{-4}$ & 6.79$\times10^{-5}$  \\  
NSSNN & 1.74$\times10^{-3}$  &  9.81$\times10^{-2}$ &  1.98$\times10^{-1}$ \\        
G-SympNet & 1.05$\times10^{-3}$  &  1.47$\times10^{-1}$  &  7.88$\times10^{-1}$  \\ 
LA-SympNet & 6.92$\times10^{-4}$  &  1.48$\times10^{-1}$ &  1.96$\times10^{-1}$ \\ 
$\alpha$-SGHN & 5.58$\times10^{-1}$  & 4.82$\times10^{-1}$   &   6.06$\times10^{+1}$ \\ 
  HGIN (Ours) & $\mathbf{2.76\times10^{-7}}$  &  $\mathbf{1.21\times10^{-7}}$ &  $\mathbf{2.01\times10^{-7}}$    \\           
\end{tabular}
\end{ruledtabular}
\end{table*}

﻿
﻿

Table~\ref{tab:model_DNLS2} and Fig.~\ref{fig_kglri_state}\,B  compare the prediction performance of HGIN and baseline models in non-separable system with heterogeneous node dynamics. HGIN outperforms all baselines in both generalization and long-term prediction accuracy.

For training details and more comparative and ablation experiments, please refer to the Appendix.

\section{Conclusions and Future Challenges}\label{sec:conclusions}
 
We have introduced HGIN, a method that simultaneously uncovers the interaction structure of a lattice Hamiltonian system from trajectory data and predicts its evolution. The approach applies to both separable and non-separable Hamiltonians, and to both homogeneous and heterogeneous node dynamics---two regimes in which previous graph-based methods either require the topology to be given or fail to converge. The structure-learning module recovers the interaction graph directly from data, and the subgraph-partitioned trajectory-prediction module assigns each cluster its own encoder, circumventing the parameter-sharing bottleneck that prevents standard GNNs from representing heterogeneous dynamics. We validated HGIN on three benchmark lattices---a Klein--Gordon chain with long-range couplings and two discrete nonlinear Schr\"odinger lattices (homogeneous and heterogeneous)---and found that it outperforms MLP, HNN, SympNet, NSSNN, and $\alpha$-SGHN in prediction accuracy, long-time energy stability, and generalization, surpassing these baseline models by several orders of magnitude in each case.

It would be particularly interesting to seek to characterize experimental
settings, e.g., from data arising in optical systems, toward classifying their
potential connectivity and also toward having a predictive ability in 
association with their observed dynamics. Additionally, while our methodology
was presented in one-dimensional settings, it is well-known~\cite{kevrekidis2009discrete,pgkreview2}
that such models are particularly relevant (and have been experimentally explored~\cite{Flach2008}) in higher dimensional settings. As such, generalizations
of the present methodology to networks related to higher-dimensional couplings would
be especially interesting and challenging to consider. Such efforts are
presently in progress and will be reported in future publications.

\paragraph*{Code and data availability.} An implementation of HGIN together with scripts to regenerate the KG-LRI and DNLS benchmarks reported in Sec.~III will be made available at a public repository upon publication.

\section*{Acknowledgments}
R. Geng was partially supported by  NSFC grant  [Grant No. 12501703]. 
This research was partly conducted while P.G.K. was 
visiting Seoul National University ---whose hospitality he acknowledges---
under the auspices of a Fulbright fellowship.
This work was also supported by the Simons Foundation
[SFI-MPS-SFM-00011048, P.G.K.]
Y.X. Gao was supported by the Science and Technology Development Plan Project of Jilin Province [Grant Nos. 20240101006JJ] and NSFC grants [Grant Nos. 12371187].
 H.-K. Zhang was partially supported by  NSFC grant  [Grant No. 12571196] and Guangdong Basic and Applied Basic Research Foundation  [Grant No. 2025A1515011952].
 J. Zu was partially supported by the Science and Technology
Development Plan Project of Jilin Province [Grant Nos. 20250102016JC].

\section*{Competing interests}

The authors declare that they have no conflict of interest.

\appendix

 \section{Data generation}
Consider a system of $N$ particles. Trajectories are obtained by numerically integrating the equations of motion with the fourth-order Runge--Kutta method (RK4) over the interval $[0,10]$\,s, with a time step of $0.001$\,s and an absolute tolerance of $10^{-12}$. The initial condition (IC) was:
 \begin{eqnarray*}
&&q_0(n)=\lambda_{n} \sin\bigg(\frac{(n-1) \pi}{N-1}\bigg),\quad n=1,\cdots,N,\\
&&p_0(n)=0,\quad n=1,\cdots,N,
\end{eqnarray*}
where $\lambda_{n} \sim \mathcal{N}(0,1)$ and $ \mathcal{N}$ denotes the standard normal distribution. The boundary condition (BC) was periodic, given by $q_{i+N}=q_i$.
We downsampled the computed trajectories at a fixed interval of 0.05s to form the dataset.
 
 \section{Model parameters}

According to the grid search method, the optimal network parameter settings for the baseline model are shown in Table \ref{tab:kglri-par}, \ref{tab:dnls_par}, and \ref{tab:dnls2_par} for the three test examples.

 \begin{table*}[]
 \centering
\caption{ For the KG-LRI system, the optimal parameters of the baseline model.
}
\label{tab:kglri-par}
\renewcommand{\arraystretch}{1.3} 
\begin{ruledtabular}
\begin{tabular}{lcccc }
\multicolumn{1}{c}{}   & Hidden layer  &  Width &  Activation function                         & Learning rate\\
\midrule
MLP  & 1 & 600 & Tanh & $1 \times 10^{-3}$ \\
HNN & 1 &  400 & SiLu & $1 \times 10^{-4}$ \\  
NSSNN & 5 &  100 &  Sigmoid & $1 \times 10^{-3}$\\        
G-SympNet & 10    & 5 & Sigmoid  & $1 \times 10^{-4}$\\ 
LA-SympNet & 5 &  4 &  Sigmoid & $1 \times 10^{-4}$\\          
\end{tabular}
\end{ruledtabular}
\end{table*}

 \begin{table*}[]
 \centering
\caption{ For the DNLS system, the optimal parameters of the baseline model.
}
\label{tab:dnls_par}
\renewcommand{\arraystretch}{1.3} 
\begin{ruledtabular}
\begin{tabular}{lcccc }
\multicolumn{1}{c}{}   & Hidden layer  &  Width &  Activation function                         & Learning rate\\
\midrule
MLP  & 1 & 200 & SiLu & $1 \times 10^{-3}$ \\
HNN & 1 &  200 & Tanh & $1 \times 10^{-4}$(3000) $1 \times 10^{-5}$\\  
NSSNN & 5 &  64 &  Sigmoid & $1 \times 10^{-4}$(3000) $1 \times 10^{-5}$\\        
G-SympNet & 20    & 10 & Sigmoid  & $1 \times 10^{-3}$\\ 
LA-SympNet & 20 &  4 &  Sigmoid & $1 \times 10^{-4}$\\          
\end{tabular}
\end{ruledtabular}
\end{table*}

 \begin{table*}[]
 \centering
\caption{ For the DNLS system, the optimal parameters of the baseline model with the second case of heterogeneous node dynamics.
}
\label{tab:dnls2_par}
\renewcommand{\arraystretch}{1.3} 
\begin{ruledtabular}
\begin{tabular}{lcccc }
\multicolumn{1}{c}{}   & Hidden layer  &  Width &  Activation function                         & Learning rate\\
\midrule
MLP  & 1 & 600 & SiLu & $1 \times 10^{-3}$ (3000)  $1 \times 10^{-4}$ \\
HNN & 2 &  800 & Tanh & $1 \times 10^{-4}$(3000) $1 \times 10^{-5}$\\  
NSSNN & 5 &  200 &  Sigmoid & $1 \times 10^{-4}$(3000) $1 \times 10^{-5}$\\        
G-SympNet & 10    & 2 & Sigmoid  & $1 \times 10^{-4}$\\ 
LA-SympNet & 20 &  4 &  Sigmoid & $1 \times 10^{-4}$\\          
\end{tabular}
\end{ruledtabular}
\end{table*}

The same network parameters are used for the three test sets of HGIN. 
The encoder $g_{noc}$ and $g_{enc}$ employ a three-layer fully-connected structure with 100 hidden units per layer, using the tanh activation function.  
The hyperparameter $\gamma$ is set to 0.05 and the learning rate is $10^{-3}$.
The total epoch is 2000 for structure learning module, and the batch size is 256.
For the trajectory prediction part, the training settings of HGIN and baseline are the same: the total number of epochs is 10,000, and the learning rate scheduling adopts a piecewise constant decay strategy \cite{montavon2012neural}. The initial learning rate is $10^{-3}$, decays to $10^{-4}$ at the 3,000th epoch, and the batch size is 256.

\section{Comparison of structural-learning modules}
In this section, we take KG-LRI as an example to demonstrate the importance of the structural learning module configuration through comparative experiments.

 \subsection{Replacement of the graph learning loss}
 In this subsection, we test the importance of the $\mathcal{L}_{GL}$ setting. Fig. \ref{fig_app_com1} presents a comparison of the system's weighted adjacency matrices learned under different $\mathcal{L}_{GL}$ settings. Clearly, only our method learns weights that are consistent with the ground truth.
\begin{figure*}[htp]  %
\centering
\includegraphics[scale=0.6]{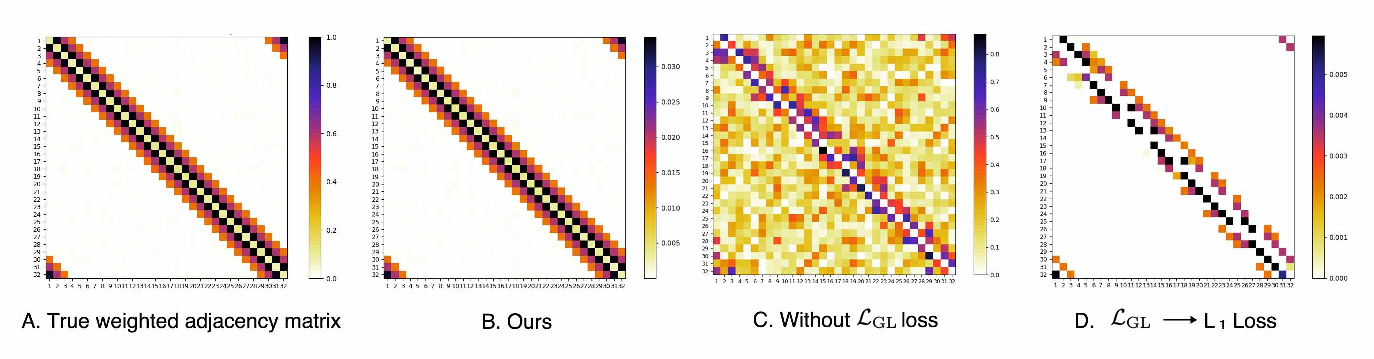}
\caption{ Comparing the importance of $\mathcal{L}_{GL}$ settings.
A. ground-truth weighted adjacency matrix of the system.
B. weighted adjacency matrix learned by our method.
C. weighted adjacency matrix learned without $\mathcal{L}_{GL}$.
D. weighted adjacency matrix learned  $\mathcal{L}_{GL}$ replaced by $L_1$ loss.
}\label{fig_app_com1}
\end{figure*} 

\subsection{ Comparison of weighted adjacency matrix module configurations}
In this subsection, we test the importance of the adjacency matrix weight learning setup. We replace the learning approach for the weights \( w_{ij} \) in our model with Graph Attention Neural Network (GAT) \cite{velivckovic2017graph}, Neural Relational Inference (NRI) \cite{kipf2018neural}, and Transformer architecture  \cite{vaswani2017attention}, respectively, and present the learned results in Fig. \ref{fig_app_com2}. Clearly, only our method yields the correct results.

\begin{figure*}[htp]  %
\centering
\includegraphics[scale=0.5]{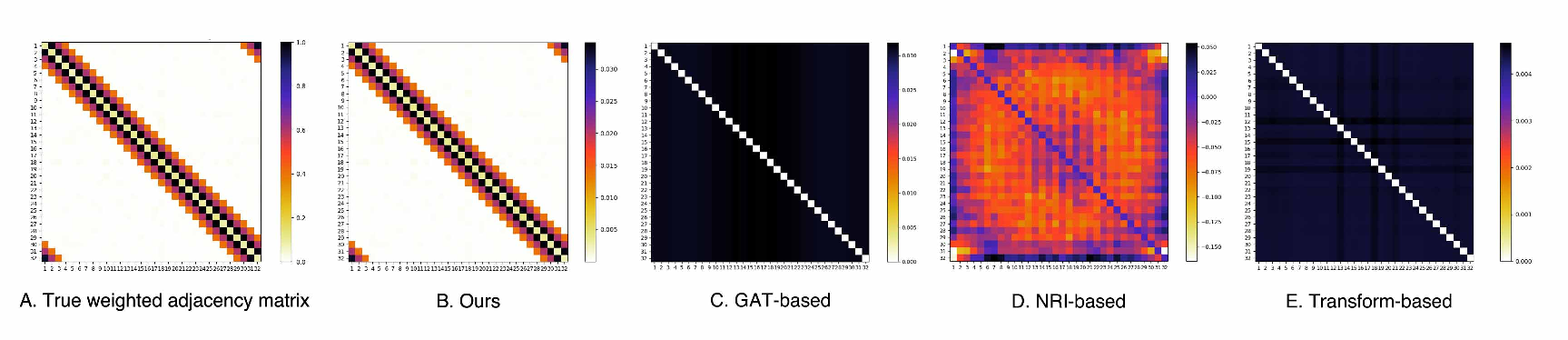}
\caption{ Comparing the importance of learning approach for the weights \( w_{ij} \).
A. ground-truth weighted adjacency matrix of the system.
B. weighted adjacency matrix learned by our method.
C. weighted adjacency matrix learned by GAT-based.
D. weighted adjacency matrix learned by NRI-based.
E. weighted adjacency matrix learned by Transformer-based.
}\label{fig_app_com2}
\end{figure*} 
\subsection{ Comparison of structural learning methods}
 In this section, we compare our structure learning method with those of Equivariant Multi-agent Motion Prediction (EqMotion) \cite{xu2023eqmotion} and NRI. The learned weighted adjacency matrices are shown in Fig. \ref{fig_app_com3}. Clearly, only our method correctly captures the interactions and their intensity relationships within the system.

\begin{figure*}[htp]  %
\centering
\includegraphics[scale=0.6]{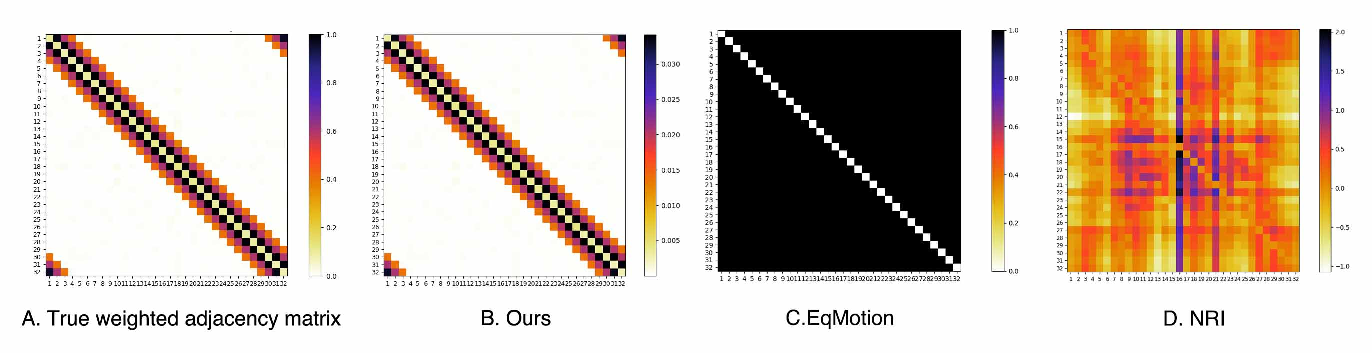}
\caption{Comparison of structural learning methods. A. ground-truth weighted adjacency matrix of the system.
B. weighted adjacency matrix learned by our method.
C. weighted adjacency matrix learned by EqMotion.
D. weighted adjacency matrix learned by NRI.
}\label{fig_app_com3}
\end{figure*}

\section{ The importance of subgraph learning}

To isolate the contribution of the subgraph learning strategy, we train an ablation variant which we call HGIN-oracle.
In HGIN-oracle, the weighted adjacency matrix $\mathcal{W}_{\theta}$ is fixed. To be more precise, we set the parts clustered as 0 in the matrix to zero to exclude machine errors. This matrix is then substituted into Equation (3), where only the parameters of the encoder are optimized, and gamma in Equation (4) is set to 0. This is equivalent to the prediction model adopting a trained attention mechanism. Still using KG-LRI as an example, Table \ref{tab:oracle_kglri} shows the prediction performance of HGIN-oracle and HGIN. Although HGIN-oracle employs approximately accurate weights, errors accumulate continuously during long-term prediction, making this method unsuitable for long-term forecasting.

 \begin{table*}[]
 \centering
\caption{KG-LRI benchmark: train loss, test loss, and energy MSE in 20s (twice the training horizon), averaged over 30 trajectories.}
\label{tab:oracle_kglri}
\renewcommand{\arraystretch}{1.3} %
\begin{ruledtabular}
\begin{tabular}{lccc }
\multicolumn{1}{c}{}   & Train loss  & Test loss & Energy MSE                         \\
\midrule
 HGIN-oracle & 3.14$\times10^{-7}$   &  3.14$\times10^{-7}$   &   9.77$\times10^{-4}$  \\ 
  HGIN & 
  $\mathbf{ 2.03 \times 10^{-8}}$  & 
  $\mathbf{ 2.39 \times 10^{-8}}$  &  
  $\mathbf{ 1.53 \times 10^{-7}}$ \\     
\end{tabular}
\end{ruledtabular}
\end{table*}

\section{Noise experiment}

To verify the robustness of the model, we added Gaussian noise with a standard deviation of $\sigma^2=0.005$, $\sigma^2=0.01$ and $\sigma^2=0.02$ on KG-LRI benchmark.  
 Fig. \ref{fig_noise_kglri} and  Table .\ref{tab:noise_kglri} respectively show the results of structure learning and trajectory prediction after adding noise. As the noise increases, the method still exhibits a certain degree of robustness.

 \begin{figure*}[htp]  %
\centering
\includegraphics[scale=0.5]{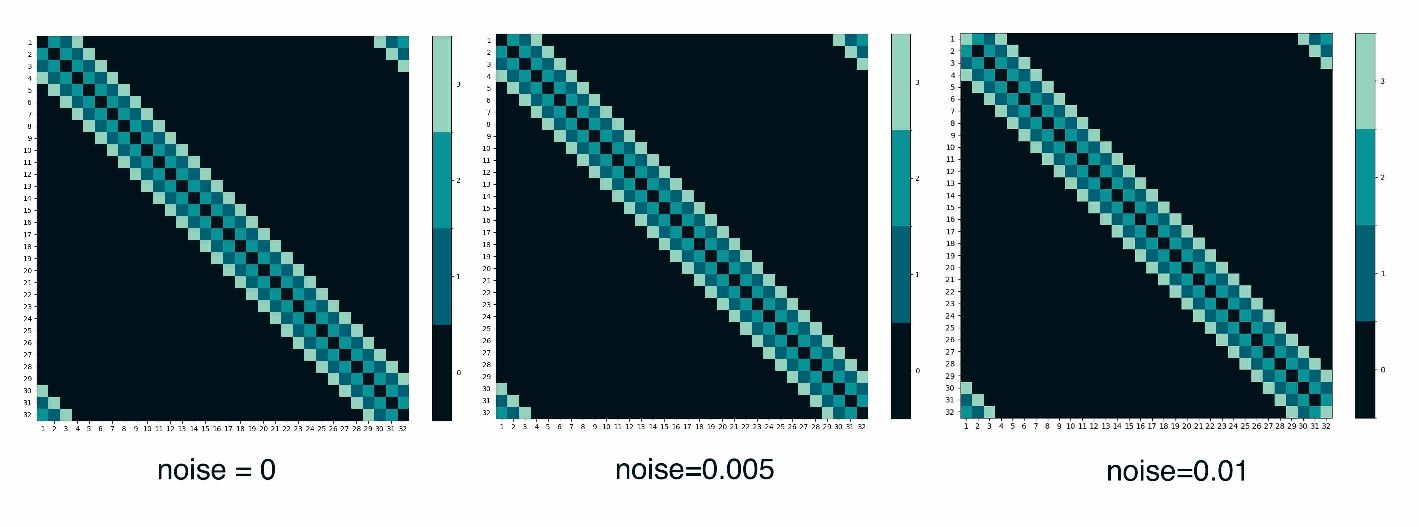}
\caption{Noise experiment on KG-LRI benchmark. The three subfigures respectively show the classification of the weight adjacency matrix output by the structure learning part after adding noise levels of 0, 0.005, and 0.001.
}\label{fig_noise_kglri}
\end{figure*} 

 \begin{table*}[]
 \centering
\caption{Noise experiment on KG-LRI benchmark: test loss, and energy MSE in 20s (twice the training horizon), averaged over 30 trajectories.}
\label{tab:noise_kglri}
\begin{ruledtabular}
\begin{tabular}{lccc }
\multicolumn{1}{c}{}  Noise & Test loss & Energy MSE    & Trajectory MSE                      \\
\midrule
0 & 
  $ 2.39 \times 10^{-8}$  &  
  $ 1.53 \times 10^{-7}$ &
     $ 2.63 \times 10^{-7}$\\
  0.005 &  
  $8.76 \times 10^{-8}$  &  
  $ 7.44 \times 10^{-7}$
  & $3.32 \times 10^{-6}$\\ 
   0.01&  
  $ 1.80 \times 10^{-7}$  &  
  $2.64 \times 10^{-6}$ &
   $9.88 \times 10^{-6}$ \\    
\end{tabular}
\end{ruledtabular}
\end{table*}

\normalem
\bibliography{apssamp}

@article{lewenstein2007ultracold,
  title={Ultracold atomic gases in optical lattices: mimicking condensed matter physics and beyond},
  author={Lewenstein, Maciej and Sanpera, Anna and Ahufinger, Veronica and Damski, Bogdan and Sen, Aditi and Sen, Ujjwal},
  doi={10.1080/00018730701223200},
  journal={Advances in Physics},
  volume={56},
  number={2},
  pages={243--379},
  year={2007},
  publisher={Taylor \& Francis}
}

@article{sai2017optimal,
  title={Optimal k-means clustering method using silhouette coefficient},
  author={Sai, L Nitya and Shreya, M Sai and Subudhi, A Anjan and Lakshmi, B Jaya and Madhuri, KB},
  doi={10.5958/0975-8089.2017.00030.6},
  journal={Int J Appl Res Inf Technol Comput},
  volume={8},
  pages={335},
  year={2017}
}

@inproceedings{kipf2018neural,
  title={Neural relational inference for interacting systems},
  author={Kipf, Thomas and Fetaya, Ethan and Wang, Kuan-Chieh and Welling, Max and Zemel, Richard},
  doi={10.48550/arXiv.1802.04687},
  booktitle={International Conference on Machine Learning},
  pages={2688--2697},
  year={2018},
  organization={PMLR}
}

@article{vaswani2017attention,
  title={Attention is all you need},
  author={Vaswani, Ashish and Shazeer, Noam and Parmar, Niki and Uszkoreit, Jakob and Jones, Llion and Gomez, Aidan N and Kaiser, {\L}ukasz and Polosukhin, Illia},
  journal={Advances in Neural Information Processing Systems},
  volume={30},
  year={2017}
}

@inproceedings{xu2023eqmotion,
  title={Eqmotion: Equivariant multi-agent motion prediction with invariant interaction reasoning},
  author={Xu, Chenxin and Tan, Robby T and Tan, Yuhong and Chen, Siheng and Wang, Yu Guang and Wang, Xinchao and Wang, Yanfeng},
  booktitle={Proceedings of the IEEE/CVF Conference on Computer Vision and Pattern Recognition},
  pages={1410--1420},
  year={2023}
}

@article{hu2019novel,
  title={A novel interval three-way concept lattice model with its application in medical diagnosis},
  author={Hu, Junhua and Chen, Dan and Liang, Pei},
  doi={10.3390/math7010103},
  journal={Mathematics},
  volume={7},
  number={1},
  pages={103},
  year={2019},
  publisher={MDPI}
}

@article{pgkreview2,
    author = {Kevrekidis, P. G.},
    title = {Non-linear waves in lattices: past, present, future},
    doi={10.1093/imamat/hxr015},
    journal = {IMA Journal of Applied Mathematics},
    volume = {76},
    number = {3},
    pages = {389-423},
    year = {2011},
    month = {04},
    issn = {0272-4960},
}

@article{chriseil,
  title = {The Discrete Nonlinear {S}chr\"odinger equation - 20 Years on},
  author = {J. C. Eilbeck and M. Johansson},
  doi={10.1142/9789812704627_0003},
  journal = {Localization and Energy Transfer in Nonlinear Systems},
  pages = {pp. 44-67},
  year = {2003}
}

@article{Flach2008,
author = {Flach, Sergej and Gorbach, Andrey V.},
doi = {10.1016/j.physrep.2008.05.002},
isbn = {0370-1573},
issn = {03701573},
journal = {Physics Reports},
month = {oct},
number = {1-3},
pages = {1--116},
title = {{Discrete breathers — Advances in theory and applications}},
volume = {467},
year = {2008}
}

@article{bpk,
  author    = {Steven L. Brunton and Joshua L. Proctor and J. Nathan Kutz},
  title     = {Discovering governing equations from data by sparse identification of nonlinear dynamical systems},
  journal   = {Proceedings of the National Academy of Sciences},
  year      = {2016},
  volume    = {113},
  number    = {15},
  pages     = {3932--3937},
  doi       = {10.1073/pnas.1517384113},
}

@article{rudy2017data,
  title={Data-driven discovery of partial differential equations},
  author={Rudy, Samuel H and Brunton, Steven L and Proctor, Joshua L and Kutz, J Nathan},
  doi={10.1038/s41467-025-65114-2},
  journal={Science Advances},
  volume={3},
  number={4},
  pages={e1602614},
  year={2017},
  publisher={AAAS}
}

@article{karniadakis2021physics,
  title={Physics-informed machine learning},
  author={Karniadakis, George Em and Kevrekidis, Ioannis G and Lu, Lu and Perdikaris, Paris and Wang, Sifan and Yang, Liu},
  doi={10.1038/s42254-021-00314-5},
  journal={Nature Reviews Physics},
  volume={3},
  number={6},
  pages={422--440},
  year={2021},
  publisher={Nature Publishing Group}
}

@misc{PGKreview,
      title={Machine Learning of Nonlinear Waves: Data-Driven Methods for Computer-Assisted Discovery of Equations, Symmetries, Conservation Laws, and Integrability}, 
      author={Jimmie Adriazola and Panayotis G. Kevrekidis and Vassilis Koukouloyannis and Wei Zhu},
      doi={10.48550/arXiv.2510.15069},
      year={2025},
      eprint={2510.15069},
      archivePrefix={arXiv},
}

@inproceedings{
xiong2021nonseparable,
title={Nonseparable Symplectic Neural Networks},
author={Shiying Xiong and Yunjin Tong and Xingzhe He and Shuqi Yang and Cheng Yang and Bo Zhu},
doi={10.48550/arXiv.2010.12636},
booktitle={International Conference on Learning Representations},
year={2021},
}

@article{ramakrishna2023bio,
  title={{Bio-inspired 3D-printed lattice structures for energy absorption applications: A review}},
  author={Ramakrishna, Doodi and Bala Murali, Gunji},
  doi={10.1177/14644207221121948},
  journal={Proceedings of the Institution of Mechanical Engineers, Part L: Journal of Materials: Design and Applications},
  volume={237},
  number={3},
  pages={503--542},
  year={2023},
  publisher={SAGE Publications Sage UK: London, England}
}

@article{coe2019lattice,
  title={Lattice density-functional theory for quantum chemistry},
  author={Coe, Jeremy P},
  doi={10.1103/PhysRevB.99.165118},
  journal={Physical Review B},
  volume={99},
  number={16},
  pages={165118},
  year={2019},
  publisher={APS}
}

@article{zok2019integrating,
  title={Integrating lattice materials science into the traditional processing--structure--properties paradigm},
  author={Zok, Frank W},
  doi={10.1557/mrc.2019.152},
  journal={Mrs Communications},
  volume={9},
  number={4},
  pages={1284--1291},
  year={2019},
  publisher={Cambridge University Press}
}

@article{bruneton2012dynamics,
  title={Dynamics of a lattice {Universe}: the dust approximation in cosmology},
  doi={10.1088/0264-9381/29/15/155001},
  author={Bruneton, Jean-Philippe and Larena, Julien},
  journal={Classical and Quantum Gravity},
  volume={29},
  number={15},
  pages={155001},
  year={2012},
  publisher={IOP Publishing}
}

@article{jiao2015pepx,
  title={{PEPX}-type lattice design and optimization for the High Energy Photon Source},
  doi={10.1088/1674-1137/39/6/067004},
  author={Jiao, Yi and Xu, Gang},
  journal={Chinese Physics C},
  volume={39},
  number={6},
  pages={067004},
  year={2015},
  publisher={IOP Publishing}
}

@article{gu2019link,
  title={Link prediction via graph attention network},
  author={Gu, Weiwei and Gao, Fei and Lou, Xiaodan and Zhang, Jiang},
  journal={arXiv preprint arXiv:1910.04807},
  doi={ 10.48550/arXiv.1910.04807},
  year={2019},
}

@article{gao2025alpha,
  title={$\alpha$-separable graph Hamiltonian network: A robust model for learning particle interactions in lattice systems},
  author={Gao, Yixian and Geng, Ru and Kevrekidis, Panayotis and Zhang, Hong-Kun and Zu, Jian},
  doi={10.1103/PhysRevE.111.015309},
  journal={Physical Review E},
  volume={111},
  number={1},
  pages={015309},
  year={2025},
  publisher={APS}
}

@article{atkinson2022improved,
  title={Improved constraints on effective top quark interactions using edge convolution networks},
  author={Atkinson, Oliver and Bhardwaj, Akanksha and Brown, Stephen and Englert, Christoph and Miller, David J and Stylianou, Panagiotis},
  doi={10.1007/JHEP04(2022)137},
  journal={Journal of High Energy Physics},
  volume={2022},
  number={4},
  pages={1--20},
  year={2022},
  publisher={Springer}
}

@article{liang2020cryspnet,
  title={{CRYSPNet}: Crystal structure predictions via neural networks},
  author={Liang, Haotong and Stanev, Valentin and Kusne, A Gilad and Takeuchi, Ichiro},
  doi={10.1103/PhysRevMaterials.4.123802},
  journal={Physical Review Materials},
  volume={4},
  number={12},
  pages={123802},
  year={2020},
  publisher={APS}
}

@article{schuetz2022combinatorial,
  title={Combinatorial optimization with physics-inspired graph neural networks},
  author={Schuetz, Martin JA and Brubaker, J Kyle and Katzgraber, Helmut G},
  doi={10.1038/s42256-022-00468-6},
  journal={Nature Machine Intelligence},
  volume={4},
  number={4},
  pages={367--377},
  year={2022},
  publisher={Nature Publishing Group UK London}
}

@inproceedings{bishnoi2023learning,
  title={Learning the dynamics of physical systems with hamiltonian graph neural networks},
  author={Bishnoi, Suresh and Bhattoo, Ravinder and Jayadeva, Jayadeva and Ranu, Sayan and Krishnan, NM Anoop},
  booktitle={ICLR 2023 Workshop on Physics for Machine Learning},
  year={2023}
}

@article{robinson2022physics,
  title={Physics guided neural networks for modelling of non-linear dynamics},
  author={Robinson, Haakon and Pawar, Suraj and Rasheed, Adil and San, Omer},
  doi={10.1016/j.neunet.2022.07.023},
  journal={Neural Networks},
  volume={154},
  pages={333--345},
  year={2022},
  publisher={Elsevier}
}

@article{chen2022deep,
  title={Deep neural network modeling of unknown partial differential equations in nodal space},
  author={Chen, Zhen and Churchill, Victor and Wu, Kailiang and Xiu, Dongbin},
  doi={10.1016/j.jcp.2021.110782},
  journal={Journal of Computational Physics},
  volume={449},
  pages={110782},
  year={2022},
  publisher={Elsevier}
}

@article{cao2024laplace,
  title={Laplace neural operator for solving differential equations},
  author={Cao, Qianying and Goswami, Somdatta and Karniadakis, George Em},
  doi={10.1038/s42256-024-00844-4},
  journal={Nature Machine Intelligence},
  volume={6},
  number={6},
  pages={631--640},
  year={2024},
  publisher={Nature Publishing Group UK London}
}

@article{chen2025data,
  title={Data-driven discovery of conservation laws from trajectories via neural deflation},
  author={Chen, Shaoxuan and Kevrekidis, Panayotis G and Zhang, Hong-Kun and Zhu, Wei},
  doi={10.1016/j.cnsns.2024.108563},
  journal={Communications in Nonlinear Science and Numerical Simulation},
  volume={143},
  pages={108563},
  year={2025},
  publisher={Elsevier}
}

@article{saqlain2023discovering,
  title={Discovering governing equations in discrete systems using {PINNs}},
  author={Saqlain, Sheikh and Zhu, Wei and Charalampidis, Efstathios G and Kevrekidis, Panayotis G},
  doi={10.1016/j.cnsns.2023.107498},
  journal={Communications in Nonlinear Science and Numerical Simulation},
  volume={126},
  pages={107498},
  year={2023},
  publisher={Elsevier}
}

@article{liu2021machine,
  title={Machine learning conservation laws from trajectories},
  author={Liu, Ziming and Tegmark, Max},
  doi={10.1103/PhysRevLett.126.180604},
  journal={Physical Review Letters},
  volume={126},
  number={18},
  pages={180604},
  year={2021},
  publisher={APS}
}

@article{lu2023discovering,
  title={Discovering conservation laws using optimal transport and manifold learning},
  author={Lu, Peter Y and Dangovski, Rumen and Solja{\v{c}}i{\'c}, Marin},
  doi={10.1038/s41467-023-40325-7},
  journal={Nature Communications},
  volume={14},
  number={1},
  pages={4744},
  year={2023},
  publisher={Nature Publishing Group UK London}
}

@article{ruiz2023neural,
  title={Neural {ODE} control for classification, approximation, and transport},
  author={Ruiz-Balet, Domenec and Zuazua, Enrique},
   doi={10.1137/21M1411433},
  journal={SIAM Review},
  volume={65},
  number={3},
  pages={735--773},
  year={2023},
  publisher={SIAM}
}

@article{arnold2022replacing,
  title={Replacing neural networks by optimal analytical predictors for the detection of phase transitions},
  author={Arnold, Julian and Sch{\"a}fer, Frank},
   doi={10.1103/PhysRevX.12.031044},
  journal={Physical Review X},
  volume={12},
  number={3},
  pages={031044},
  year={2022},
  publisher={APS}
}

@article{radhakrishnan2023wide,
  title={Wide and deep neural networks achieve consistency for classification},
  author={Radhakrishnan, Adityanarayanan and Belkin, Mikhail and Uhler, Caroline},
   doi={10.1073/pnas.2208779120},
  journal={Proceedings of the National Academy of Sciences},
  volume={120},
  number={14},
  pages={e2208779120},
  year={2023},
  publisher={National Academy of Sciences}
}

@article{nuske2023finite,
  title={Finite-data error bounds for {Koopman}-based prediction and control},
  author={N{\"u}ske, Feliks and Peitz, Sebastian and Philipp, Friedrich and Schaller, Manuel and Worthmann, Karl},
  doi={10.1073/pnas.2208779120},
  journal={Journal of Nonlinear Science},
  volume={33},
  number={1},
  pages={14},
  year={2023},
  publisher={Springer}
}

@article{david2023symplectic,
  title={Symplectic learning for {Hamiltonian} neural networks},
  author={David, Marco and M{\'e}hats, Florian},
  doi={10.1016/j.jcp.2023.112495},
  journal={Journal of Computational Physics},
  volume={494},
  pages={112495},
  year={2023},
  publisher={Elsevier}
}

@book{montavon2012neural,
  title={Neural networks: tricks of the trade},
  author={Montavon, Gr{\'e}goire and Orr, Genevi{\`e}ve and M{\"u}ller, Klaus-Robert},
  doi = {10.1007/3-540-49430-8},
  volume={7700},
  year={2012},
  publisher={springer}
}

@article{jin2020sympnets,
  title={SympNets: Intrinsic structure-preserving symplectic networks for identifying {Hamiltonian} systems},
  author={Jin, Pengzhan and Zhang, Zhen and Zhu, Aiqing and Tang, Yifa and Karniadakis, George Em},
  doi={10.1016/j.neunet.2020.08.017},
  journal={Neural Networks},
  volume={132},
  pages={166--179},
  year={2020},
  publisher={Elsevier}
}

@article{efremidis2002discrete,
  title={Discrete solitons in nonlinear zigzag optical waveguide arrays with tailored diffraction properties},
  author={Efremidis, Nikos K and Christodoulides, Demetrios N},
  doi={10.1364/NLGW.2002.NLMD35},
  journal={Physical Review E},
  volume={65},
  number={5},
  pages={056607},
  year={2002},
  publisher={APS}
}

@article{greydanus2019hamiltonian,
  title={Hamiltonian neural networks},
  author={Greydanus, Samuel and Dzamba, Misko and Yosinski, Jason},
  doi = {10.48550/arXiv.1906.01563},
  journal={Advances in Neural Information Processing Systems 32 (NeurIPS 2019)},
  volume={32},
  year={2019}
}

@misc{goodfellow2016deep,
  title={Deep Learning},
  author={Goodfellow, Ian and  Bengio, Yoshua and Courville, Aaron},
  year={2016},
  publisher={MIT Press},
url={https://books. google. com/books},
}

@article{jin2022learning,
  title={Learning {Poisson} systems and trajectories of autonomous systems via {Poisson} neural networks},
  author={Jin, Pengzhan and Zhang, Zhen and Kevrekidis, Ioannis G and Karniadakis, George Em},
  doi={10.1109/TNNLS.2022.3148734},
  journal={IEEE Transactions on Neural Networks and Learning Systems},
  volume={34},
  number={11},
  pages={8271--8283},
  year={2022},
  publisher={IEEE}
}

@article{elmar,
  title = {Experimental Observation of Single- and Multisite Matter-Wave Solitons in an Optical Accordion Lattice},
  author = {Cruickshank, Robbie and Lorenzi, Francesco and La Rooij, Arthur and Kerr, Ethan F. and Hilker, Timon and Kuhr, Stefan and Salasnich, Luca and Haller, Elmar},
  journal = {Phys. Rev. Lett.},
  volume = {135},
  issue = {26},
  pages = {263404},
  numpages = {8},
  year = {2025},
  month = {Dec},
  publisher = {American Physical Society},
  doi = {10.1103/sh72-wnmv},
  url = {https://link.aps.org/doi/10.1103/sh72-wnmv}
}

@article{SzameitRechtsman2024DiscreteNonlinearTopologicalPhotonic,
  author    = {Alexander Szameit and Mikael C. Rechtsman},
  title     = {Discrete nonlinear topological photonics},
  journal   = {Nature Physics},
  year      = {2024},
  volume    = {20},
  number    = {6},
  pages     = {905--912},
  doi       = {10.1038/s41567-024-02454-8},
  url       = {https://doi.org/10.1038/s41567-024-02454-8}
}

@article{kevrekidis2009discrete,
  title={The discrete nonlinear Schr{\"o}dinger equation: mathematical analysis, numerical computations and physical perspectives},
  author={Kevrekidis, Panayotis G},
  doi={10.1007/978-3-540-89199-4},
  volume={232},
  year={2009},
  journal={Springer Science \& Business Media}
}

@article{raissi2019physics,
  title={Physics-informed neural networks: A deep learning framework for solving forward and inverse problems involving nonlinear partial differential equations},
  author={Raissi, Maziar and Perdikaris, Paris and Karniadakis, George E},
  doi = {10.1016/j.jcp.2018.10.045},
  journal={Journal of Computational Physics},
  volume={378},
  pages={686--707},
  year={2019},
  publisher={Elsevier}
}

@article{saqlain2022discovering,
  title={Discovering governing equations in discrete systems using {PINNs}},
  author={Saqlain, Sheikh and Zhu, Wei and Charalampidis, Efstathios G and Kevrekidis, Panayotis G},
  doi={10.48550/arXiv.2212.00971},
  journal={arXiv preprint arXiv:2212.00971},
  year={2022}
}

@article{sanchez2019hamiltonian,
  title={Hamiltonian graph networks with {ODE} integrators},
   doi={10.48550/arXiv.1909.12790},
  author={Sanchez-Gonzalez, Alvaro and Bapst, Victor and Cranmer, Kyle and Battaglia, Peter},
  journal={arXiv preprint arXiv:1909.12790},
  year={2019}
}

@article{velivckovic2017graph,
  title={Graph attention networks},
  author={Veli{\v{c}}kovi{\'c}, Petar and Cucurull, Guillem and Casanova, Arantxa and Romero, Adriana and Lio, Pietro and Bengio, Yoshua},
  doi = {10.48550/arXiv.1710.10903},
  journal={International Conference on Learning Representations},
  year={2017}
}

@article{zhu2022neural,
  title={Neural networks enforcing physical symmetries in nonlinear dynamical lattices: The case example of the {Ablowitz--Ladik} model},
  author={Zhu, Wei and Khademi, Wesley and Charalampidis, Efstathios G and Kevrekidis, Panayotis G},
  doi={10.1016/j.physd.2022.133264},
  journal={Physica D: Nonlinear Phenomena},
  volume={434},
  pages={133264},
  year={2022},
  publisher={Elsevier}
}

@article{zvyagintseva2022machine,
  title={Machine learning of phase transitions in nonlinear polariton lattices},
  author={Zvyagintseva, Daria and Sigurdsson, Helgi and Kozin, Valerii K and Iorsh, Ivan and Shelykh, Ivan A and Ulyantsev, Vladimir and Kyriienko, Oleksandr},
  doi={10.1038/s42005-021-00755-5},
  journal={Communications Physics},
  volume={5},
  number={1},
  pages={8},
  year={2022},
  publisher={Nature Publishing Group UK London}
}

\end{document}